%% file: cvpr.tex
\documentclass[final]{cvpr}

\usepackage{times}
\usepackage{epsfig}
\usepackage{graphicx}
\usepackage{amsmath}
\usepackage{amssymb}

\usepackage{graphicx}
\usepackage{comment}
\usepackage{color}
\usepackage{microtype}
\usepackage{subfigure}
\usepackage{booktabs} 
\usepackage{amsmath}
\usepackage{amssymb}
\usepackage{caption}
\usepackage{multirow}
\usepackage[ruled,vlined]{algorithm2e}
\usepackage{setspace}
\usepackage{tabularx}
\usepackage{makecell}
\usepackage[normalem]{ulem}

\newcommand{\hs}[1]{\textcolor{black}{#1}}
\newcommand{\cm}[1]{\textcolor{black}{#1}}

\let\svmaketitle\maketitle
\def\maketitle{\svmaketitle\thispagestyle{empty}}

\makeatletter
\newcommand{\printfnsymbol}[1]{%
  \textsuperscript{\@fnsymbol{#1}}%
}
\makeatother

\usepackage[pagebackref=true,breaklinks=true,colorlinks,bookmarks=false]{hyperref}

\def\cvprPaperID{7002} 
\def\confYear{CVPR 2021}

\begin{document}

\title{Reducing Domain Gap by Reducing Style Bias}

\author{
Hyeonseob Nam\thanks{Equal contribution}\quad
HyunJae Lee\printfnsymbol{1}\quad
Jongchan Park\quad
Wonjun Yoon\quad
Donggeun Yoo\\
Lunit Inc.\\
{\tt\small \{hsnam,hjlee,jcpark,wonjun,dgyoo\}@lunit.io}
}

\maketitle

\input{0.abstract}

\input{1.introduction}

\input{2.related_work}

\input{3.sagnets}

\input{4.experiments}

\input{5.conclusion}

{\small
\bibliographystyle{ieee_fullname}
\bibliography{cvpr}
}

\end{document}

%% file: 0.abstract.tex
\begin{abstract}
Convolutional Neural Networks (CNNs) often fail to maintain their performance when they confront new test domains, which is known as the problem of domain shift.
Recent studies suggest that one of the main causes of this problem is CNNs' strong inductive bias towards image styles (i.e. textures) which are sensitive to domain changes, rather than contents (i.e. shapes).
Inspired by this, we propose to reduce the intrinsic style bias of CNNs to close the gap between domains.
Our Style-Agnostic Networks (SagNets) disentangle style encodings from class categories to prevent style biased predictions and focus more on the contents.
Extensive experiments show that our method effectively reduces the style bias and makes the model more robust under domain shift.
It achieves remarkable performance improvements in a wide range of cross-domain tasks including domain generalization, unsupervised domain adaptation, and semi-supervised domain adaptation on multiple datasets.\footnote{Code: 
\url{https://github.com/hyeonseobnam/sagnet}}
\end{abstract}

%% file: 1.introduction.tex
\begin{figure*}[t]
\begin{center}
    \includegraphics[width=0.77\textwidth]{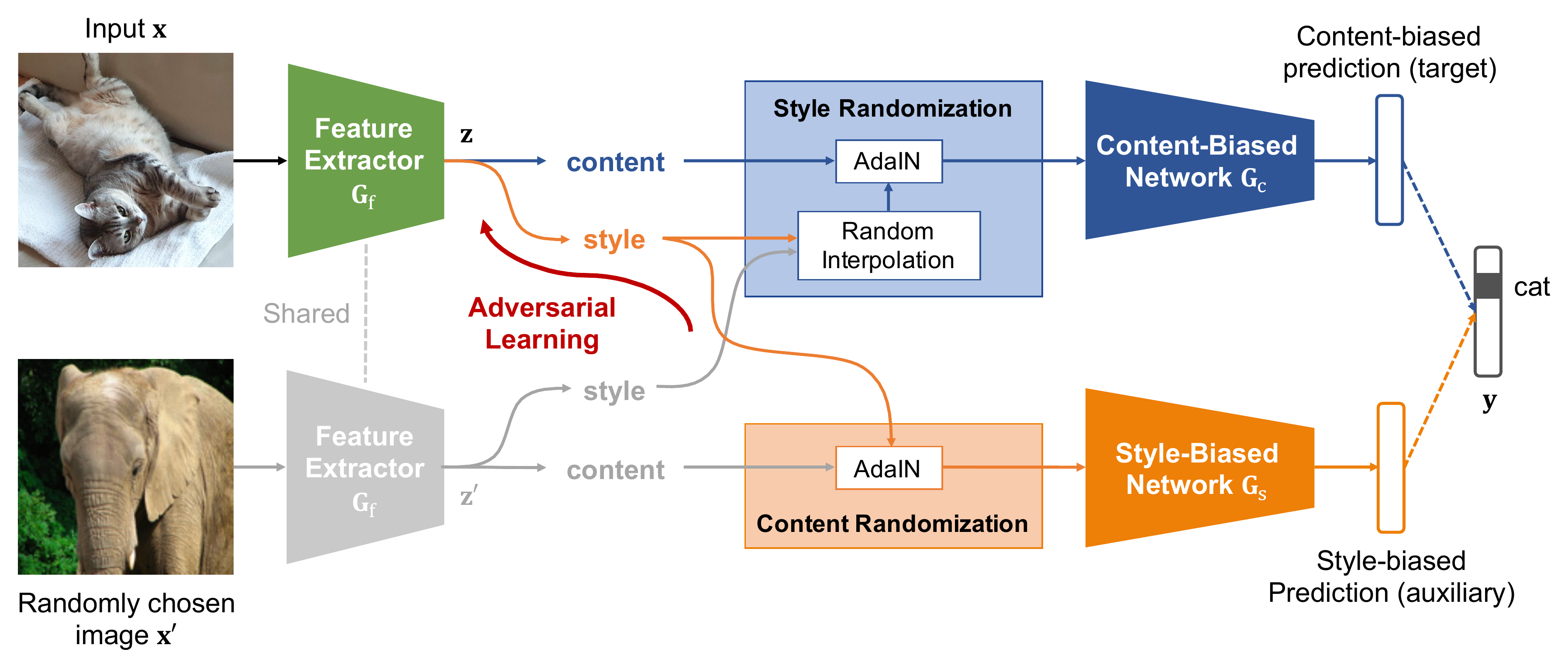}
    \caption{
    Our Style-Agnostic Network (SagNet) reduces style bias to reduce domain gap.
    It consists of three sub-networks---a feature extractor, a content-biased network, and a style-biased network---that are jointly trained end-to-end.
    The content-biased network is led to focus on the content of the input via style randomization (SR), where the style of the input is replaced by an arbitrary style through AdaIN~\cite{huang2017arbitrary}.
    Conversely, the style-biased network is led to focus on the style by content randomization (CR), while an adversarial learning makes the feature extractor generate less style-biased representation.
    }
    \label{fig:sagnet}
    \vspace{-2mm}
\end{center}
\end{figure*}

\section{Introduction}
\label{sec:introduction}

Despite the huge success of Convolutional Neural Networks (CNNs) fueled by large-scale training data, their performance often degrades significantly when they encounter test data from unseen environments.
This phenomenon, known as the problem of domain shift~\cite{pan2009survey}, comes from the representation gap between training and testing domains.
For a more reliable deployment of CNNs to ever-changing real-world scenarios, the community has long sought to make CNNs robust to domain shift
under various problem settings such as Domain Generalization (DG)~\cite{li2017deeper,innocente2018domain,carlucci2019domain,dou2019domain}, Unsupervised Domain Adaptation (UDA)~\cite{mingsheng2015learning,ganin2016domain,tzeng2017adversarial,mingsheng2018conditional,peng2019moment}, and Semi-Supervised Domain Adaptation (SSDA)~\cite{donahue2013semi,yao2015semi,saito2019semi}.

In stark contrast to the vulnerability of CNNs against domain shift, the human visual recognition system generalizes incredibly well across domains.
For example, young children learn many object concepts from pictures, but they naturally transfer their knowledge to the real world~\cite{ganea2008transfer}.
Similarly, people can easily recognize objects in cartoons or paintings even if they have not seen the same style of an image before. 
Where does such a difference come from?

A recent line of studies has revealed that standard CNNs have an inductive bias far different from human vision: while humans tend to recognize objects based on their contents (i.e. shapes)~\cite{landau1988importance}, CNNs exhibit a strong bias towards styles (i.e. textures)~\cite{Baker2018DeepCN,geirhos2019imagenet,Hermann2019ExploringTO}.
This may explain why CNNs are intrinsically more sensitive to domain shift because image styles are more likely to change across domains than the contents.
Geirhos et al.~\cite{geirhos2019imagenet} supported this hypothesis by showing that CNNs trained with heavy augmentation on styles become more robust against various image distortions.
Research on CNN architectures~\cite{nam2018batch,lee2019srm} has also demonstrated that adjusting the style information in CNNs helps to address multi-domain tasks.

In this paper, we experimentally analyze the relation between CNNs' inductive bias and representation gap across domains, and exploit this relation to address domain shift problems.
We propose \textit{Style-Agnostic Networks (SagNets)} which 
effectively improve CNNs' domain transferability by controlling their inductive bias, without directly reducing domain discrepancy.
Our framework consists of separate content-biased and style-biased networks on top of a feature extractor.
The content-biased network is encouraged to focus on contents by randomizing styles in a latent space.
The style-biased network is led to focus on styles in the opposite way, against which the feature extractor adversarially makes the styles incapable of discriminating class categories.
At test time, the prediction is made by the combination of the feature extractor and the content-biased network, where the style bias is substantially reduced.

We show that there exists an apparent correlation between CNNs' inductive bias and their ability to handle domain shift: reducing style bias reduces domain discrepancy.
Based on this property, SagNets make significant improvements in a wide range of domain shift scenarios including DG, UDA, and SSDA, across several cross-domain benchmarks such as PACS~\cite{li2017deeper}, Office-Home~\cite{venkateswara2017deep}, and DomainNet~\cite{peng2019moment}.

Our method is orthogonal to the majority of existing domain adaptation and generalization techniques that utilize domain information for training (e.g. aligning the source and target domains~\cite{ganin2016domain,mingsheng2018conditional,saito2019semi}).
In other words, SagNets only control the intrinsic bias of CNNs without even requiring domain labels nor multiple domains.
This approach is not only scalable to more practical scenarios where domain boundaries are unknown or ambiguous, but also able to complement existing methods and bring additional performance boosts as demonstrated in our extensive experiments.

%% file: 2.related_work.tex
\section{Related Work}
\label{sec:related_work}

\paragraph{\textnormal{\textbf{\hs{Inductive biases} of CNNs.}}}
Our work is motivated by the recent findings that CNNs tend to learn styles rather than contents.
Geirhos et al.~\cite{geirhos2019imagenet} observed that standard ImageNet-trained CNNs are likely to make a style-biased decision on ambiguous stimuli (e.g. images stylized to different categories).
Some studies have also shown that CNNs perform well when only local textures are given while global shape structures are missing~\cite{gatys2015texture,brendel2019bagnets}, but work poorly in the reverse scenario~\cite{Ballester2016OnTP,geirhos2019imagenet}.
Others have attempted to mitigate the style bias of CNNs under the assumption that it deteriorates CNNs' generalization capability~\cite{Hosseini2018AssessingSB,geirhos2019imagenet,Hermann2019ExploringTO}.
However, the practical impact of CNNs' bias on domain shift problems remains unclear, which we aim to explore in this paper.

\paragraph{\textnormal{\textbf{Style Manipulation.}}}
We utilize convolutional feature statistics to control the style bias of CNNs.
This owes to previous works dealing with the feature statistics in CNNs, mostly in generative frameworks to manipulate image styles.
Gatys et al.~\cite{gatys2015texture} showed that feature statistics of a CNN effectively capture the style information of an image, which paved the way for neural style transfer~\cite{gatys2016image,johnson2016perceptual,huang2017arbitrary}.
In particular, adaptive instance normalization (AdaIN)~\cite{huang2017arbitrary} demonstrated that the style of an image can easily be changed by adjusting the mean and variance of convolutional feature maps.
StyleGAN~\cite{karras2019style} also produced impressive image generation results by repeatedly applying AdaIN operations in a generative network.
Manipulating styles has also benefited discriminative problems.
BIN~\cite{nam2018batch} improved classification performance by reducing unnecessary style information using trainable normalization, and SRM~\cite{lee2019srm} extended this idea to style-based feature recalibration.

\paragraph{\textnormal{\textbf{Domain Generalization and Adaptation.}}}
Domain Generalization (DG) aims to make CNNs robust against novel domains outside the training distribution.
A popular approach is learning a shared feature space across multiple source domains, for example, by minimizing Maximum Mean Discrepancy (MMD)~\cite{muandet2013domain,ghifary2016scatter} or adversarial feature alignment~\cite{li2018domain,li2018deep}.
Some studies divide the model into domain-specific and domain-invariant components using low-rank parameterization~\cite{li2017deeper} or layer aggregation modules~\cite{innocente2018domain}, assuming that the domain-invariant part well generalizes to other domains.
Meta-learning frameworks~\cite{balaji2018metareg,dou2019domain} split the source domains into meta-train and meta-test domains to simulate domain shift, and JiGen~\cite{carlucci2019domain} employed self-supervised signals such as solving jigsaw puzzles to improve generalization by learning image regularities.

Unsupervised Domain Adaptation (UDA) tackles the problem of domain shift where unlabeled data from the target domain are available for training.
The mainstream approach is aligning the source and the target distributions by minimizing MMD~\cite{ghifary2014domain,long2017deep}, adversarial learning~\cite{ganin2016domain,tzeng2017adversarial,mingsheng2018conditional}, or image-level translation~\cite{hoffman2017cycada,murez2018image}.
Another problem setting with domain shift is Semi-Supervised Domain Adaptation (SSDA) where a few target labels are additionally provided.
It is commonly addressed by simultaneously minimizing the distance between the domains as well as imposing a regularization constraint on the target data to prevent overfitting~\cite{donahue2013semi,yao2015semi}.
Recently, MME~\cite{saito2019semi} optimized a minimax loss on the conditional entropy of unlabeled data and achieved large performance improvement.

%% file: 3.sagnets.tex
\section{Style-Agnostic Networks}
\label{sec:sagnets}
We present a general framework to make the network agnostic to styles but focused more on contents \hs{in an end-to-end manner} (Fig.~\ref{fig:sagnet}).
It contains three networks: a feature extractor, a content-biased network, and a style-biased network.
The content-biased network is encouraged to exploit image contents when making decisions by randomizing intermediate styles, which we call \textit{content-biased learning}.
Conversely, the style-biased network is trained to be biased towards styles by randomizing contents, but it adversarially makes the feature extractor less style-biased, which is referred to as \textit{adversarial style-biased learning}.
The final prediction at test time is made by the feature extractor followed by the content-biased network.
Following the common practice~\cite{gatys2016image,johnson2016perceptual,huang2017arbitrary,karras2019style,nam2018batch,lee2019srm}, we utilize the summary statistics of CNN features (i.e. channel-wise mean and standard deviation) as style representation and their spatial configuration as content representation.

\subsection{Content-Biased Learning}
In content-biased learning, we enforce the model to learn content-biased features by introducing a style randomization (SR) module.
It randomizes the styles during training by interpolating the feature statistics between different examples regardless of their class categories.
Consequently, the network is encouraged to be invariant to styles and biased toward contents in predicting the class labels.

Given an input training image $\mathbf{x}$ and a randomly selected image $\mathbf{x}'$, we first extract their intermediate feature maps $\mathbf{z}, \mathbf{z}' \in \mathbb{R}^{D \times H\times W }$ from the feature extractor $\mathbf{G}_\text{f}$, where $H$ and $W$ indicate spatial dimensions, and $D$ is the number of channels. Then we compute the channel-wise mean and standard deviation $\mu(\mathbf{z}), \sigma(\mathbf{z}) \in \mathbb{R}^D$ as style representation:
\begin{align}
\mu(\mathbf{z}) &= \frac 1 {HW} \sum_{h=1}^H \sum_{w=1}^W \mathbf{z}_{hw}, \\
\sigma(\mathbf{z}) &= \sqrt{ \frac 1 {HW} \sum_{h=1}^H \sum_{w=1}^W (\mathbf{z}_{hw} - \mu(\mathbf{z}))^2 + \epsilon}.
\end{align}
The SR module constructs a randomized style $\hat{\mu}, \hat{\sigma} \in \mathbb{R}^D$ by interpolating between the styles of $\mathbf{z}$ and $\mathbf{z}'$.
Then it replaces the style of the input with the randomized style through adaptive instance normalization (AdaIN)~\cite{huang2017arbitrary}:
\begin{align}
\hat{\mu} &= \alpha \cdot \mu(\mathbf{z}) + (1 - \alpha) \cdot \mu(\mathbf{z}'),
\\
\hat{\sigma} &= \alpha \cdot \sigma(\mathbf{z}) + (1 - \alpha) \cdot \sigma(\mathbf{z}'),
\\
\operatorname{SR}(\mathbf{z}, \mathbf{z}') &= \hat{\sigma} \cdot \left( \frac {\mathbf{z} - \mu(\mathbf{z})} {\sigma(\mathbf{z})}\right) + \hat{\mu}, \label{eq:SR}
\end{align}
where $\alpha\sim\text{Uniform}(0,1)$ is a random interpolation weight.
The style-randomized representation $\operatorname{SR}(\mathbf{z}, \mathbf{z}')$ is fed into the content-biased network $\mathbf{G}_\text{c}$ to obtain a content-biased loss $L_\text{c}$, which jointly optimizes the feature extractor $\mathbf{G}_\text{f}$ and the content-biased network $\mathbf{G}_\text{c}$:
\begin{align}
\min \limits_{\mathbf{G}_\text{f}, \mathbf{G}_\text{c}} 
L_\text{c} &= -\mathbb{E}_{(\mathbf{x}, \mathbf{y}) \in S}  \sum_{k=1}^K \mathbf{y}_k \log \mathbf{G}_\text{c}(\operatorname{SR}(\mathbf{G}_\text{f}(\mathbf{x}), \mathbf{z}'))_k, \label{eq:Lc}
\end{align}
where $K$ is the number of class categories, $\mathbf{y}\in\{0,1\}^K$ is the one-hot label for input $\mathbf{x}$ and $S$ is the training set.

By randomizing styles during training, the content-biased network can no longer rely on the style but focus more on the content in making a decision.
At test time, we remove the SR module and directly connect $\mathbf{G}_\text{f}$ to $\mathbf{G}_\text{c}$, which ensures independent predictions as well as saves computations.

\subsection{Adversarial Style-Biased Learning}
In addition to the content-biased learning,
we constrain the feature extractor from learning style-biased representation by adopting an adversarial learning framework.
In other words, we make the styles encoded by the feature extractor incapable of discriminating the class categories.
To achieve this, we build an auxiliary style-biased network $\mathbf{G}_\text{s}$ as a discriminator to make style-biased predictions utilizing a content randomization (CR) module.
Then we let the feature extractor $\mathbf{G}_\text{f}$ fool $\mathbf{G}_\text{s}$ via an adversarial framework, which we name adversarial style-biased learning.

Contrary to SR which leaves the content of the input and randomizes its style, the CR module does the opposite: it maintains the style and switches the content to that of an arbitrary image. 
Given intermediate feature maps $\mathbf{z}, \mathbf{z}' \in \mathbb{R}^{D \times H\times W }$ corresponding to the input $\mathbf{x}$ and the randomly chosen image $\mathbf{x}'$, we apply AdaIN to the content of $\mathbf{z}'$ with the style of $\mathbf{z}$:
\begin{align}
\operatorname{CR}(\mathbf{z}, \mathbf{z}') = {\sigma}(\mathbf{z}) \cdot \left( \frac {\mathbf{z}' - \mu(\mathbf{z}')} {\sigma(\mathbf{z}')}\right) + \mu(\mathbf{z}), \label{eq:CR}
\end{align}
which can be interpreted as a content-randomized representation of $
\mathbf{z}$.
It is taken as an input to the style-biased network $\mathbf{G}_\text{s}$ which is trained to make a style-biased prediction by minimizing a style-biased loss $L_\text{s}$: 
\begin{align}
\min \limits_{\mathbf{G}_\text{s}} 
L_\text{s} = -\mathbb{E}_{(\mathbf{x}, \mathbf{y}) \in S}  \sum_{k=1}^K \mathbf{y}_k \log \mathbf{G}_\text{s}(\operatorname{CR}(\mathbf{G}_\text{f}(\mathbf{x}), \mathbf{z}'))_k. \label{eq:Ls}
\end{align}
The feature extractor $\mathbf{G}_\text{f}$ is then trained to fool $\mathbf{G}_\text{s}$ by minimizing an adversarial loss $L_\text{adv}$ computed by the cross-entropy between the style-biased prediction and uniform distribution\footnote{This can also be interpreted as maximizing the entropy of style-biased predictions. We empirically found that matching with a uniform distribution provides more stable convergence than directly maximizing the entropy or using a gradient reversal layer~\cite{ganin2016domain}.}: 
\begin{align}
\min \limits_{\mathbf{G}_\text{f}} 
L_\text{adv} = -\lambda_\text{adv} \mathbb{E}_{(\mathbf{x}, \cdot) \in S}  \sum_{k=1}^K \frac{1}{K}  \log \mathbf{G}_\text{s}(\operatorname{CR}(\mathbf{G}_\text{f}(\mathbf{x}), \mathbf{z}')))_k, \label{eq:Ladv}
\end{align}
where $\lambda_\text{adv}$ is a weight coefficient.

\hs{Following previous work that utilizes affine transformation parameters (i.e. normalization parameters) to effectively manipulate the style representation~\cite{huang2017arbitrary,karras2019style,nam2018batch}, we perform the adversarial learning with respect to the affine transformation parameters of $\mathbf{G}_\text{f}$.}
We can also efficiently control the trade-off between the content and style biases by adjusting the coefficient $\lambda_\text{adv}$, which is explored in Sec.~\ref{sec:exp-bias}.

\subsection{Implementation Details}

\begin{algorithm}[t]
\setstretch{1.1}
\SetAlgoLined
\textbf{Input:} training data $S = {(\mathbf{x}_i, \mathbf{y}_i)}_{i=1}^{M}$; 
batch size $N$;
hyperparameter $\lambda_\text{adv} > 0$
\\
\textbf{Initialize:} feature extractor $\mathbf{G}_\text{f}$; content-biased network $\mathbf{G}_\text{c}$; style-biased network $\mathbf{G}_\text{s}$ 
\\
\While{not converged}{
	$\mathbf{X}, \mathbf{Y} = \textsc{SampleBatch}(S, N) $ 
	\\
    $\mathbf{Z} = \mathbf{G}_\text{f}(\mathbf{X})$ 
    \\
    $\mathbf{Z}' = \textsc{Shuffle}(\mathbf{Z})$ \tcp*[f]{\hs{randomly shuffle Z along the batch dimension.}} 
    \\
    \textbf{Content-Biased Learning:} 
    \\
    $\mathbf{Z}^{(\text{c})} = \operatorname{SR}(\mathbf{Z}, \mathbf{Z}')$ 
    \\
    $L_\text{c} = - \frac  {1} {N} \sum_{j=1}^N \sum_{k=1}^K \mathbf{Y}_{jk}\log \mathbf{G}_\text{c}(\mathbf{Z}^{(\text{c})}_j)_k$ 
    \\
    Minimize $L_\text{c}$ w.r.t. $\mathbf{G}_\text{f}$ and $\mathbf{G}_\text{c}$ 
    \\
    \textbf{Adversarial Style-Biased Learning:} 
    \\
    $\mathbf{Z}^{(\text{s})} = \operatorname{CR}(\mathbf{Z}, \mathbf{Z}')$  
    \\
    $L_\text{s} = - \frac  {1} {N} \sum_{j=1}^N \sum_{k=1}^K \mathbf{Y}_{jk}\log \mathbf{G}_\text{s}(\mathbf{Z}^{(\text{s})}_j)_k$ 
    \\  
    Minimize $L_\text{c}$ w.r.t. $\mathbf{G}_\text{s}$ 
    \\
    $L_{\text{adv}} = - \lambda_{\text{adv}}   \frac  {1} {N} \sum_{j=1}^N  \sum_{k=1}^K \frac  {1} {K} \log \mathbf{G}_\text{s}(\mathbf{Z}^{(\text{s})}_j)_k$ 
    \\ 
    Minimize $L_{\text{adv}}$ w.r.t. $\mathbf{G}_\text{f}$
}
\textbf{Output:} $\mathbf{G}_\text{c} \circ \mathbf{G}_\text{f}$
\caption{Optimization Process of SagNets}
\label{algo:sagnet}
\end{algorithm}

Our framework can be readily integrated into modern CNN architectures and trained end-to-end.
Given a CNN such as a ResNet~\cite{he2016deep}, the feature extractor $\mathbf{G}_\text{f}$ comprises first few stages\footnote{A stage refers to a group of layers that share the same feature map size.} of the CNN, while the rest of the network becomes the content-biased network $\mathbf{G}_\text{c}$; the style-biased network $\mathbf{G}_\text{s}$ forms the same structure as $\mathbf{G}_\text{c}$.
Consequently, the output network $\mathbf{G}_\text{c} \circ \mathbf{G}_\text{f}$ has exactly the same architecture as the original CNN---in other words, it does not impose any overhead in terms of both parameters and computations at test time.
To minimize the computational overhead during training, we choose the arbitrary example within a minibatch: given intermediate feature maps $\mathbf{Z} \in \mathbb{R}^{ N \times D \times H\times W}$ from a minibatch of size $N$, we construct new feature maps $\mathbf{Z}'$ corresponding to the arbitrary examples by randomly \hs{shuffling} $\mathbf{Z}$ along the batch dimension.
The overall training procedure of SagNets is summarized in Algorithm \ref{algo:sagnet}.

\begin{figure*}[t]
\begin{center}
\includegraphics[width=0.75\textwidth]{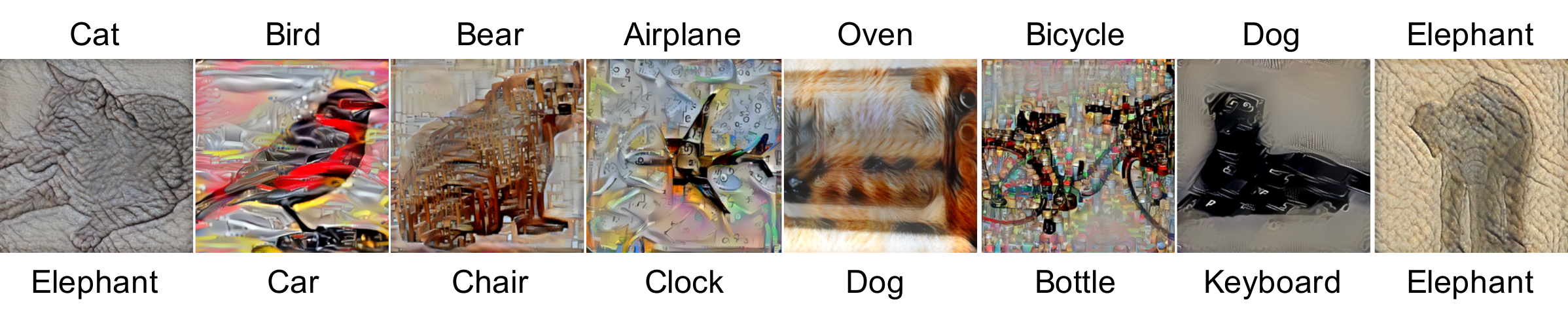}
\vspace{-1mm}
\caption{
Examples of the texture-shape cue conflict stimuli~\cite{geirhos2019imagenet} created by randomly combining the shape (top labels) and texture (bottom labels) of different images \hs{through style transfer~\cite{gatys2016image}}.
}
\label{fig:stimuli}
\vspace{-2mm}
\end{center}
\end{figure*}

\begin{figure*}[t]
\begin{center}
\
\subfigure[Texture/shape accuracy]{
\label{fig:shape-texture-acc}
\includegraphics[width=.23\linewidth]{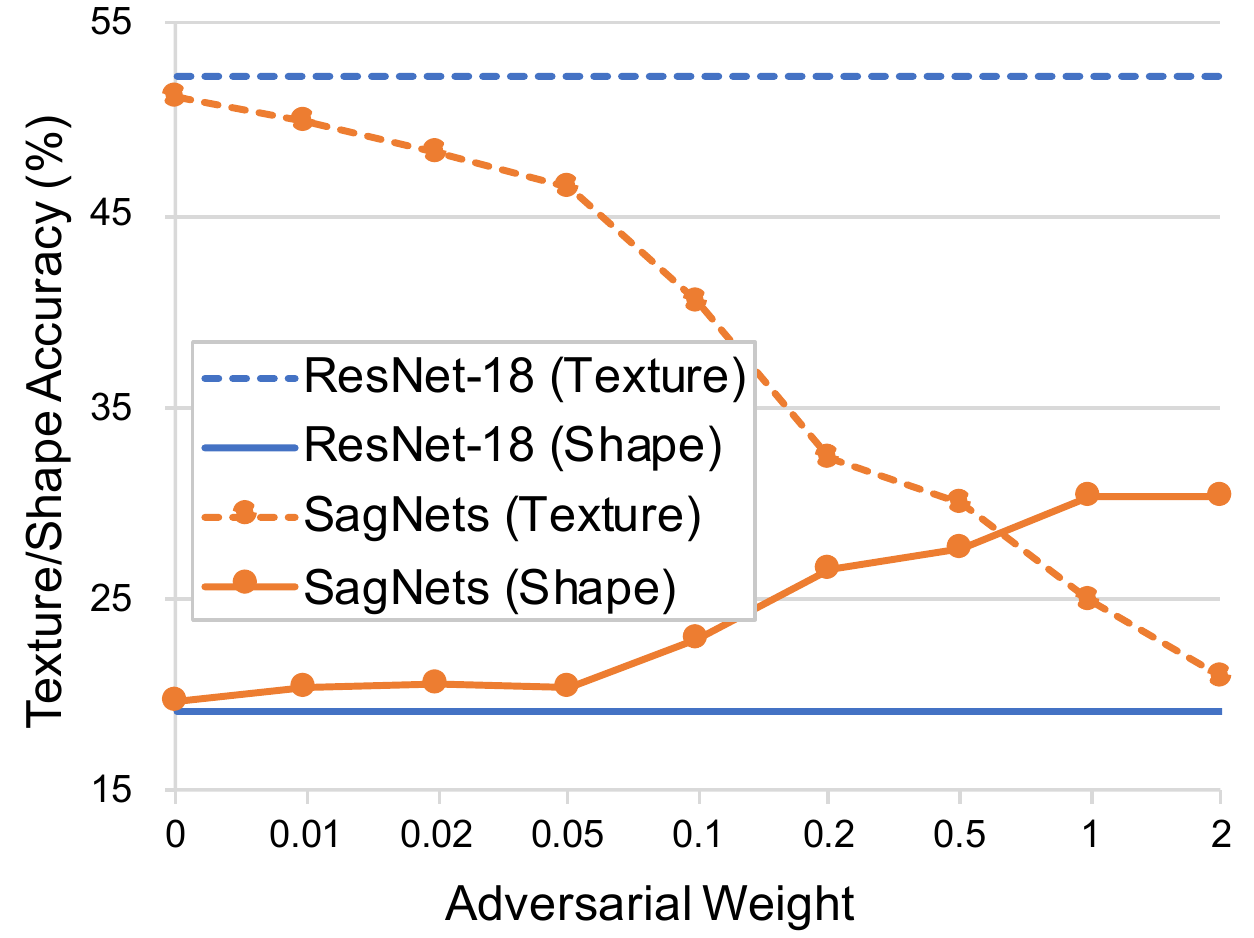}}
\
\subfigure[Shape bias]{
\label{fig:shape-texture-bias}
\includegraphics[width=.23\linewidth]{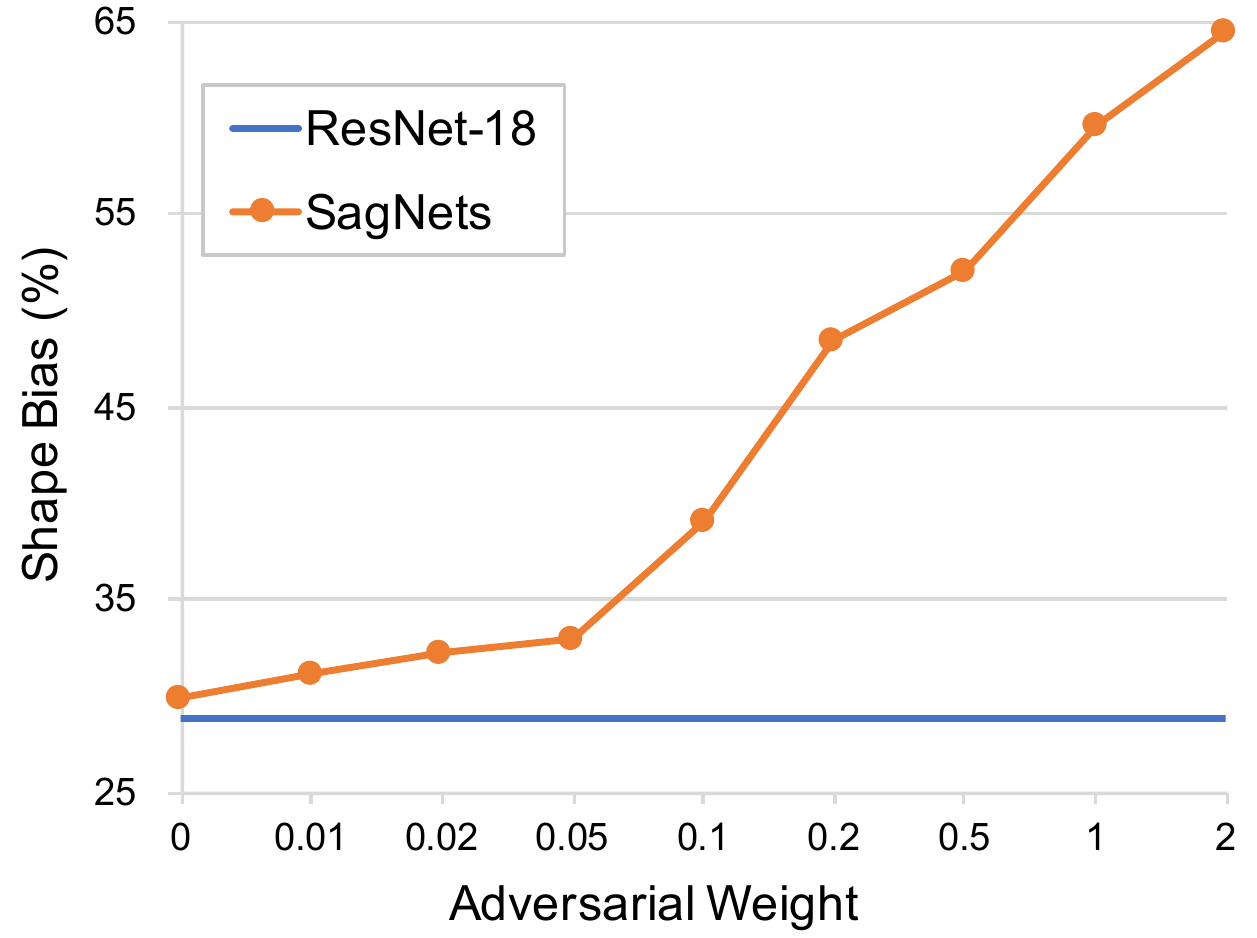}}
\
\subfigure[Domain discrepancy]{
\label{fig:adistance}
\includegraphics[width=.23\linewidth]{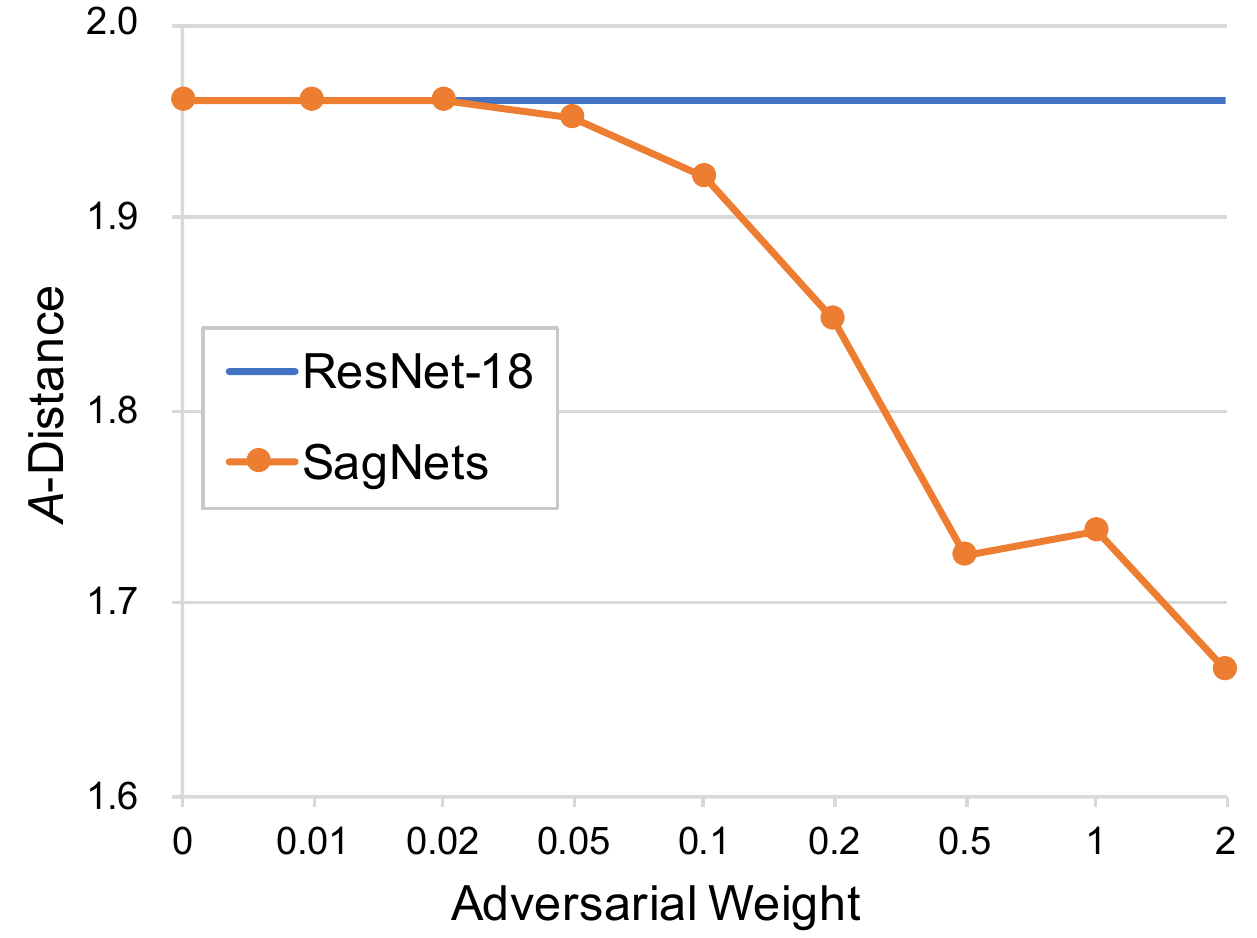}}
\
\subfigure[Domain discrepancy \textit{vs} shape bias]{
\label{fig:adistance-bias}
\includegraphics[width=0.23\linewidth]{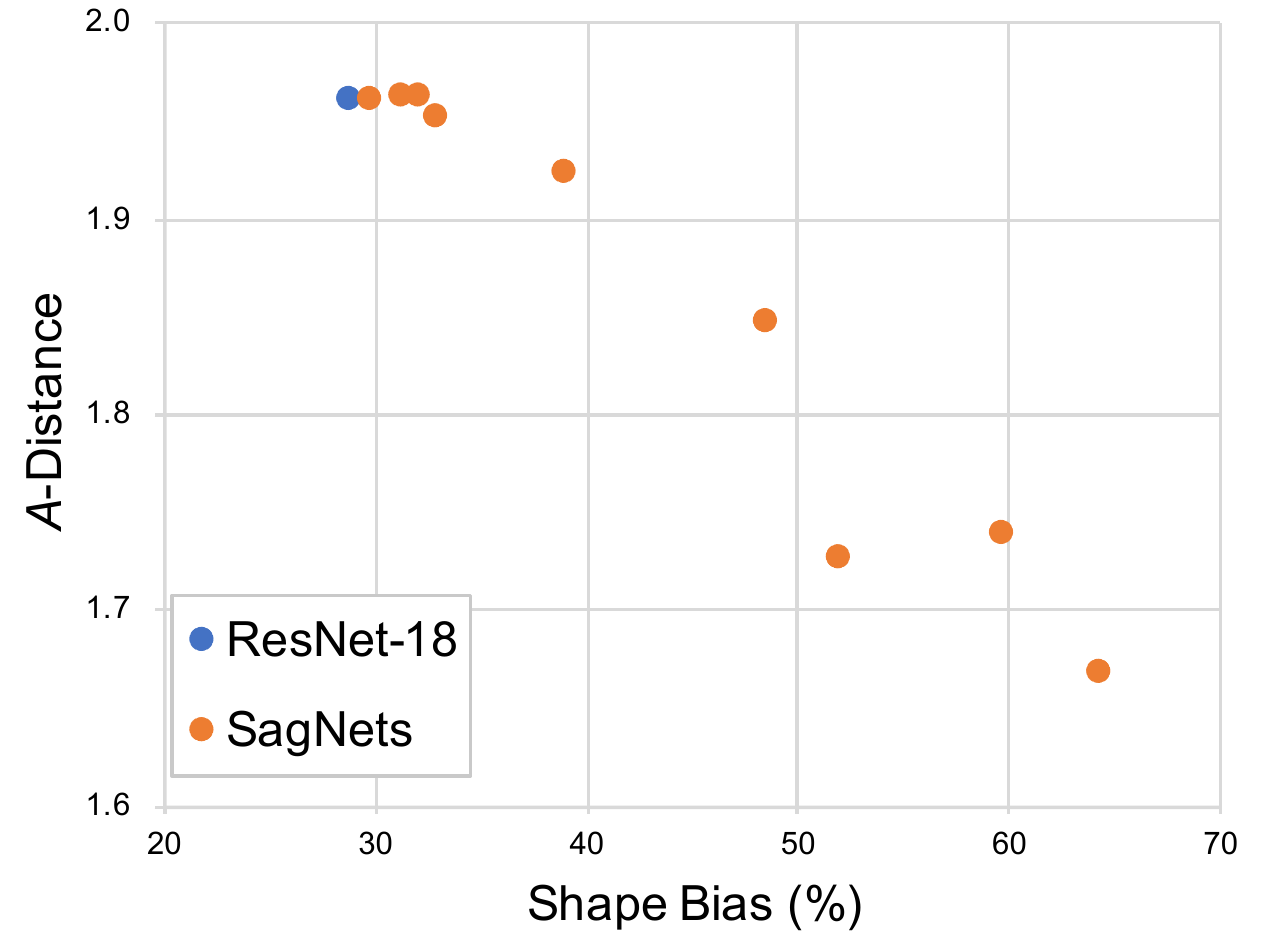}
}
\
\vspace{-1mm}
\caption{
The effect of SagNets on (a) texture/shape accuracy and (b) shape bias on the cue conflict stimuli. 
\hs{The domain discrepancy ($\mathcal{A}$-distance) between 16-class-ImageNet and the cue conflict stimuli against (c) the adversarial weight and (d) the shape bias.}
SagNets significantly increase the shape accuracy and bias, while reducing the domain discrepancy.
}
\label{fig:ab_wadv}
\vspace{-2mm}
\end{center}
\end{figure*}

\subsection{Extension to Un/Semi-Supervised Learning}
Although our framework does not require training data from the target domain, some problem settings such as un/semi-supervised domain adaptation allow access to unlabeled target data for training.
To fully leverage those unlabeled data, we utilize a simple extension of SagNets based on consistency learning~\cite{miyato2018virtual,xie2018learning}.
For each training example $\mathbf{x}$ from a set of unlabeled data $S_{\text{unl}}$, we obtain two prediction vectors from the network: one applied SR and the other not applied SR.
The inconsistency between the two predictions are estimated by the mean square error to form an unsupervised loss $L_\text{unl}$ and minimized for all unlabeled data:
\begin{multline}
\min \limits_{\mathbf{G}_\text{f}, \mathbf{G}_\text{c}} L_\text{unl} = \lambda_\text{unl} \mathbb{E}_{\mathbf{x} \in S_\text{unl}} \sum_{k=1}^K \{\mathbf{G}_\text{c}(\operatorname{SR}(\mathbf{G}_\text{f}(\mathbf{x}), \mathbf{z}'))_k
\cr
- \mathbf{G}_\text{c}(\mathbf{G}_\text{f}(\mathbf{x}))_k \}^2, \label{eq:Lunl}
\end{multline}
where $\lambda_\text{unl}$ is the corresponding coefficient which is set to 0.01.
Furthermore, our adversarial learning is naturally extended to unlabeled data because the adversarial loss (Eq.~\ref{eq:Ladv}) does not require ground-truth labels.

%% file: 4.experiments.tex
\section{Experiments}
\label{sec:experiments}
In this section, we first conduct an experimental analysis to gain an insight into the effect of SagNets (Sec.~\ref{sec:exp-bias}), then perform comprehensive evaluation on a wide range of cross-domain tasks including domain generalization (DG) (Sec.~\ref{sec:exp-dg}), unsupervised domain adaptation (UDA) (Sec.~\ref{sec:exp-uda}), and semi-supervised domain adaptation (SSDA) (Sec.~\ref{sec:exp-ssda}) compared with existing methods.
All networks are pretrained on ImageNet classification~\cite{russakovsky2015imagenet}.

\subsection{Biases and Domain Gap}
\label{sec:exp-bias}
We examine the effect of SagNets on CNNs' inductive biases and domain discrepancy using \textbf{16-class-ImageNet}~\cite{geirhos2018generalisation} and the \textbf{texture-shape cue conflict stimuli}~\cite{geirhos2019imagenet}.
16-class-ImageNet is a subset of ImageNet containing 213,555 images from 16 entry-level categories.
The texture-shape cue conflict stimuli were introduced to quantify the intrinsic biases of CNNs, which consist of 1,280 images and share the same 16 categories as 16-class-ImageNet.
They are generated by blending the texture (style) and shape (content) from different images via style transfer~\cite{gatys2016image} (see examples in Fig.~\ref{fig:stimuli}), so that we can observe whether a CNN makes a decision based on the texture or shape. 

\cm{We train SagNets with a ResNet-18 baseline on 16-class-ImageNet only, using} SGD with batch size 256, momentum 0.9, weight decay 0.0001, initial learning rate 0.001, and cosine learning rate scheduling for 30 epochs.
The randomization stage (the stage after which the randomizations are performed. i.e. the number of stages in the feature extractor) is set to 2, and we vary the adversarial coefficient $\lambda_\text{adv}$.

\paragraph{\textnormal{\textbf{Texture/Shape Bias.}}}
As proposed in \cite{geirhos2019imagenet}, we quantify the texture and shape biases of networks by evaluating them on the cue conflict stimuli and counting the number of predictions that correctly classify the texture or shape of images.
Specifically, shape bias is defined as the fraction of predictions matching the shape within the predictions matching either the shape or texture, and texture bias is defined \hs{in a similar way}.
Fig.~\ref{fig:shape-texture-acc} shows the texture/shape accuracy on the stimuli of the ResNet-18 baseline and SagNets with varying adversarial weight $\lambda_\text{adv}$, which clearly demonstrates that SagNets increase the shape accuracy and decrease the texture accuracy.
This consequently increases the shape bias as shown in Fig.~\ref{fig:shape-texture-bias}, which is also equivalent to decreasing the texture bias.
Furthermore, the shape and texture biases are effectively controlled by the adversarial weight $\lambda_\text{adv}$, i.e. the shape bias increases as $\lambda_\text{adv}$ increases.
In practice, increasing $\lambda_\text{adv}$ does not always improve the final accuracy because it makes the optimization more difficult, thus we need to find a fair trade-off, which we investigate in Sec~\ref{sec:exp-dg}.

\paragraph{\textnormal{\textbf{Domain Gap.}}}
We further investigate the capability of SagNets in reducing domain discrepancy.
We treat 16-class-ImageNet and the cue conflict stimuli dataset as two different domains because they share the same object categories but exhibit different appearances.
We then measure the distance between the two domains using the features from the penultimate layer of the network.
Following \cite{mingsheng2015learning}, we calculate a proxy $\mathcal{A}$-distance $d_\mathcal{A} = 2(1 - \epsilon)$ where $\epsilon$ is a generalization error of an SVM classifier trained to distinguish the examples from the two domains.
As illustrated in Fig.~\ref{fig:adistance}, SagNets effectively reduce the domain discrepancy as the adversarial weight increases.
By plotting the $\mathcal{A}$-distance against the shape bias as presented in Fig.~\ref{fig:adistance-bias},
we observe an explicit correlation between the bias and domain gap: shape-biased representation generalizes better across domains, which confirms the common intuition~\cite{Hosseini2018AssessingSB,geirhos2019imagenet,Hermann2019ExploringTO}.

\input{table/dg_pacs}

\input{table/dg_officehome}

\begin{figure*}[t]
\begin{center}
\
\subfigure[Accuracy \textit{vs} randomization stage]{
\label{fig:ab-sl}
\includegraphics[width=0.235\textwidth]{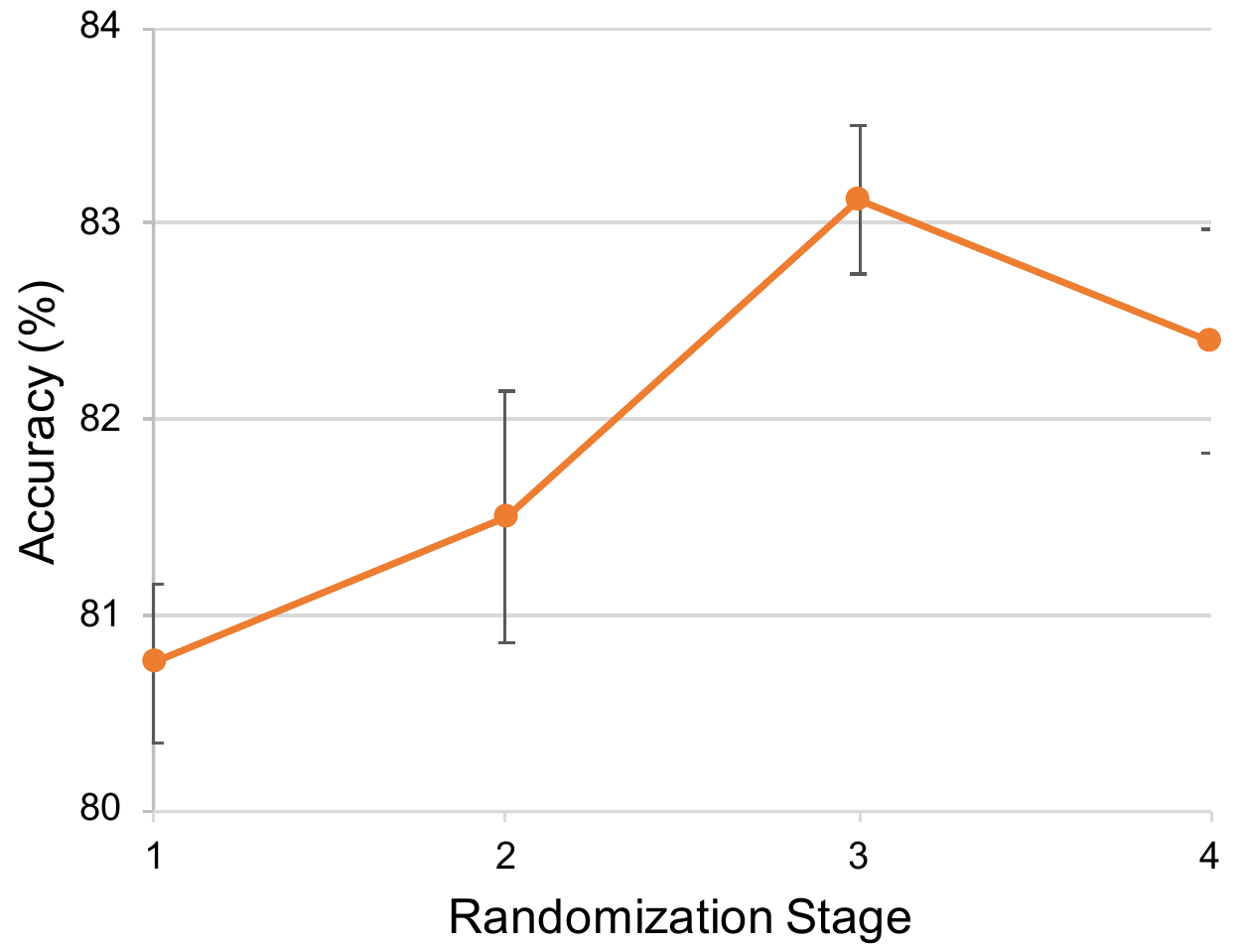}
}
\
\subfigure[Accuracy \textit{vs} adversarial weight]{
\label{fig:ab-wadv}
\includegraphics[width=0.235\textwidth]{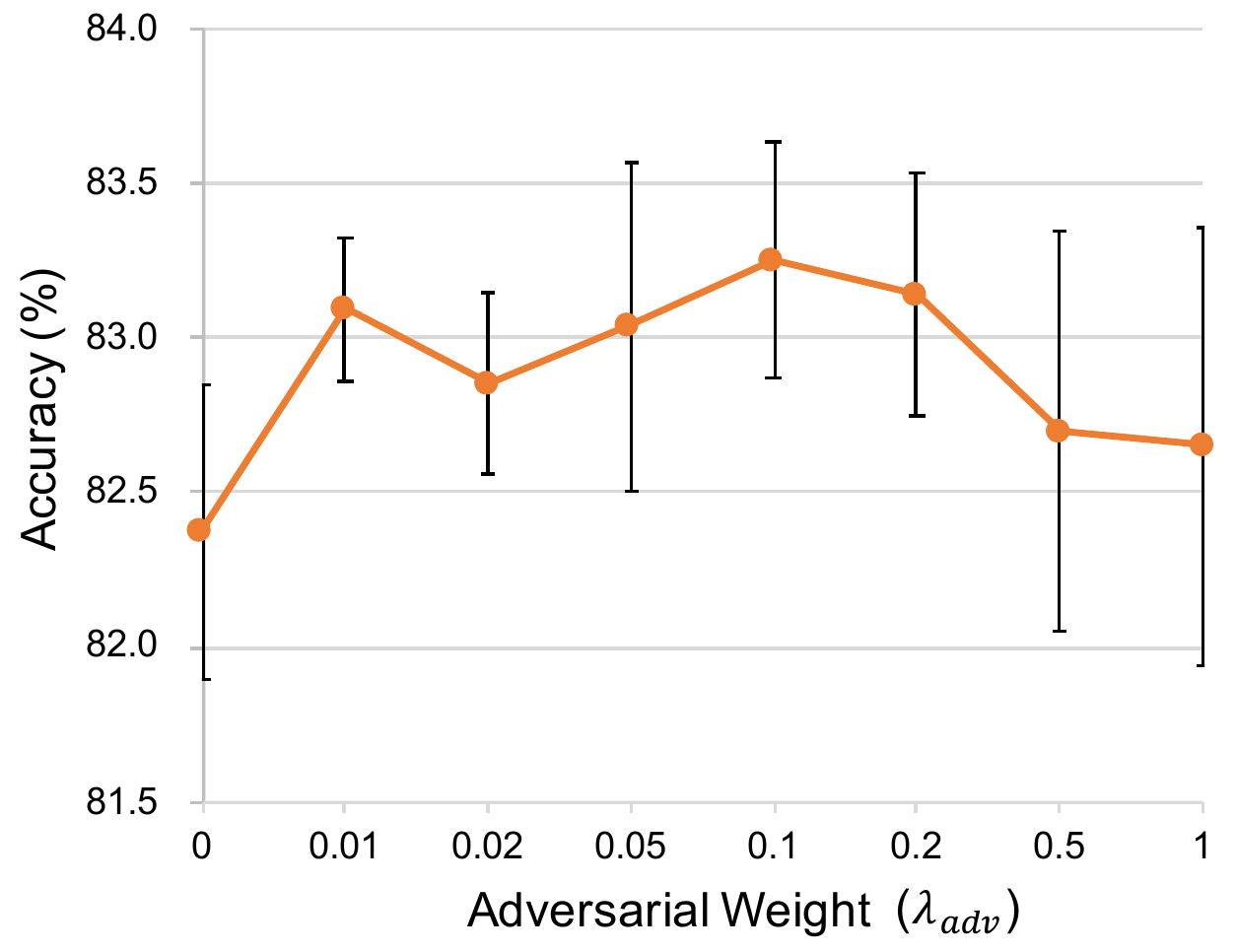}
}
\
\subfigure[Domain discrepancy]{
\label{fig:pacs-adistance}
\includegraphics[width=0.235\textwidth]{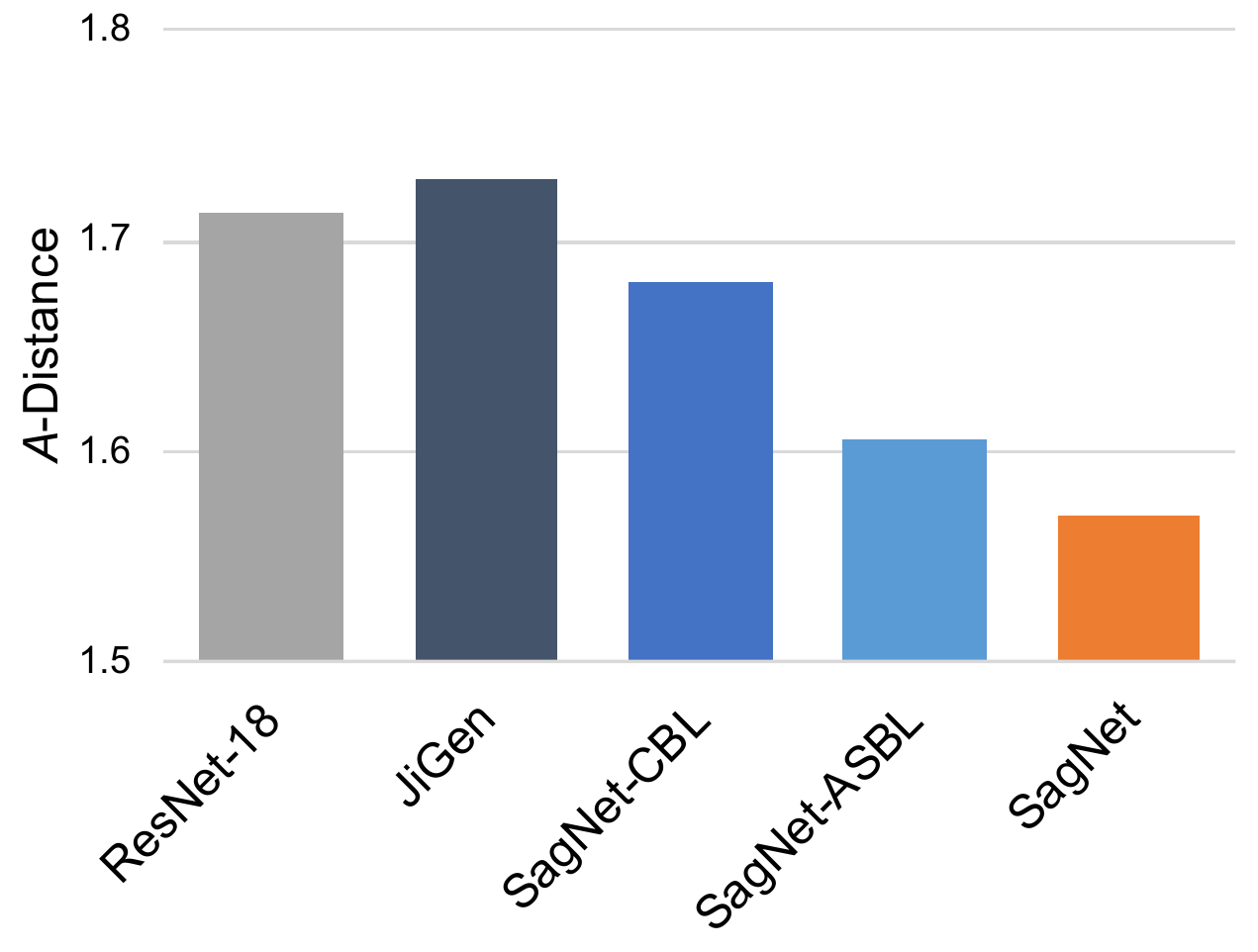}
}
\
\vspace{-1mm}
\caption{
Accuracy of SagNets on PACS with varying (a) randomization stage and (b) adversarial weight; (c) the domain discrepancy ($\mathcal{A}$-distance) between the source domains and the target domain.
Results are averaged over the 4 target domains and 3 repetitions, where error bars denote the standard deviation. 
}
\label{fig:ablation}
\vspace{-2mm}
\end{center}
\end{figure*}

\input{table/dg_pacs_single}

\input{table/uda_officehome}
\input{table/uda_domainnet_v2}
\input{table/ssda_domainnet2}

\subsection{Domain Generalization}
\label{sec:exp-dg}
DG is a problem to train a model on a single or multiple source domain(s), and test on an unseen target domain.
We evaluate the efficacy of SagNets against \hs{recent} DG methods including D-SAM~\cite{innocente2018domain}, JiGen~\cite{carlucci2019domain}, Epi-FCR~\cite{li2019episodic}, MASF~\cite{dou2019domain}, and MMLD~\cite{matsuura2020domain}. 
We adopt two multi-domain datasets: \textbf{PACS}~\cite{li2017deeper} consists of 9,991 images from 7 categories across 4 domains (Art Painting, Cartoon, Sketch, and Photo) and \textbf{Office-Home}~\cite{venkateswara2017deep} comprises 15,588 images from 65 categories and 4 domains (Art, Clipart, Product and Real-World).
We split the training data of PACS into 70\% training and 30\% validation following the official split~\cite{li2017deeper}, and Office-Home into 90\% training and 10\% validation following \cite{innocente2018domain}.
Our networks are trained by SGD with batch size 96, momentum 0.9, weight decay 0.0001, initial learning rate 0.004 (0.002 with AlexNet) and cosine scheduling for 2K iterations (4K with AlexNet or Office-Home).
The randomization stage and the adversarial weight of SagNets are fixed to 3 and 0.1, respectively, throughout all remaining experiments unless otherwise specified.

\paragraph{\textnormal{\textbf{Multi-Source Domain Generalization.}}}
We first examine multi-source DG where the model needs to generalize from multiple source domains to a novel target domain. 
We train our SagNets and the DeepAll baselines (i.e. na\"ive supervised learning) on the combination of all training data from the source domains regardless of their domain labels.
Table~\ref{table:msdg} and \ref{table:msdg-officehome} demonstrate that SagNets not only significantly improve the accuracy over the DeepAll baselines but also outperform the competing methods.
Furthermore, experiments without content-biased learning (SagNet$^{-CBL}$) and adversarial style-biases learning (SagNet$^{-ASBL}$) verify the effectiveness of each component of SagNets.
It is also worth noting that while all the compared methods except JiGen exploit additional layers on top of the baseline CNN at test time, SagNets do not require any extra parameters nor computations over the baseline.

\paragraph{\textnormal{\textbf{Ablation.}}}
We perform an extensive ablation study for SagNets on multi-source DG using PACS.
We first examine the effect of the randomization stage where the randomizations are applied, while keeping the adversarial weight to 0.1.
Karras et al.~\cite{karras2019style} demonstrated that styles at different layers encode distinct visual attributes: the styles from fine spatial resolutions (lower layers in our network) encode low-level attributes such as color and microtextures, while the styles from coarse spatial resolutions (higher layers in our network) encode high-level attributes such as global structures and macrotextures.
In this regard, the randomization modules of SagNets need to be applied at a proper level, where the style incurs undesired bias.
As shown in Fig.~\ref{fig:ab-sl}, SagNets offer the best improvement when the randomizations are applied after stage 3, 
while randomizing too low-level styles is less helpful in reducing style bias; randomizing too high-level styles may lose important semantic information.

We also conduct ablation on the adversarial coefficient $\lambda_\text{adv}$, while the randomization stage is fixed to 3.
Fig.~\ref{fig:ab-wadv} illustrates the accuracy of SagNets with varying values of $\lambda_\text{adv}$. 
Although increasing $\lambda_\text{adv}$ tends to steadily improve the shape bias as shown in Fig.~\ref{fig:shape-texture-bias}, the actual performance peaks around $\lambda_\text{adv} = 0.1$.
This indicates that while increasing the shape bias leads to reducing domain discrepancy, it may complicate the optimization and damage the performance when exceeding a proper range.

\paragraph{\textnormal{\textbf{Domain Gap.}}}
Reducing domain gap in DG setting is particularly challenging compared with other tasks such as UDA or SSDA since one can not make any use of the target distribution.
To demonstrate the effect of SagNets in such scenario, we measure the $\mathcal{A}$-distances between the source domains and the target domain following the same procedure as in Sec.~\ref{sec:exp-bias}, and average them over the 4 generalization tasks of PACS.
As shown in Fig.~\ref{fig:pacs-adistance}, SagNets considerably reduce the domain discrepancy compared to the ResNet-18 baseline, JiGen\footnote{\label{fn:jigen}We provide additional comparison with JiGen because it does not use domain labels as same as ours.
The results are reproduced with their official code (\url{https://github.com/fmcarlucci/JigenDG}) and the optimal hyperparameters provided in their paper.
}, and ablated SagNets.

\paragraph{\textnormal{\textbf{Single-Source Domain Generalization.}}}
Our framework seamlessly extends to single-source DG where only a single training domain is provided, because it does not require domain labels nor multiple source domains (which are necessary in the majority of DG methods~\cite{li2017deeper,innocente2018domain,li2019episodic,dou2019domain}).
We train SagNets on each domain of PACS and evaluate them on the remaining domains.
As reported in Table~\ref{table:ssdg}, SagNets remarkably boost the generalization performance, while JiGen and \cm{ADA\footnote{
We re-implement their method on top of our baseline, which gives better performance than reproduction based on their published code.
}~\cite{volpi2018generalizing} are technically applicable to single-source DG but outperformed by our method.}

\subsection{Unsupervised Domain Adaptation}
\label{sec:exp-uda}
UDA is a task of transferring knowledge from a source to the target domain, where unlabeled target data are available for training.
Besides \textbf{Office-Home} that we used for DG, we also employ \textbf{DomainNet}~\cite{peng2019moment} which is a large-scale dataset containing about 0.6 million images from 6 domains and 345 categories.
Since some domains and classes are very noisy, we follow \cite{saito2019semi} to utilize its subset of 145,145 images from 4 domains (Clipart, Painting, Sketch, and Real) with 126 categories, and consider 7 adaptation scenarios.
We demonstrate that SagNets are not only effective when used alone, but also able to complement popular UDA methods such as DANN~\cite{ganin2016domain}, JAN~\cite{long2017deep}, CDAN~\cite{mingsheng2018conditional}, and SymNet~\cite{zhang2019domain} to make further improvements.
\cm{
We reproduce all compared methods and their SagNet variants on top of the same baseline and training policy for fair comparison.}

As shown in Table~\ref{table:uda-officehome}, SagNets lead to impressive performance improvements when combined with any adaptation method.
Furthermore, while CDAN does not make meaningful improvement on DomainNet due to the complexity of the dataset in terms of scale and diversity, SagNets consistently improve the accuracy in most adaptation scenarios as presented in Table~\ref{table:uda-domainnet}.
These results indicate that the effectiveness of SagNets is orthogonal to existing approaches and generates a synergy by complementing them.

\subsection{Semi-Supervised Domain Adaptation}
\label{sec:exp-ssda} 
We also conduct SSDA on \textbf{DomainNet} where a few target labels are provided.
Saito et al.~\cite{saito2019semi} showed that common UDA methods~\cite{ganin2016domain,mingsheng2018conditional,saito2017adversarial} often fail to improve their performance in such scenario, and proposed minimax entropy (MME) setting the state-of-the-art for SSDA.
Here we show that SagNets are successfully applied to SSDA, bringing further improvements over MME.
For a fair comparison, SagNets are built upon the official implementation of MME\footnote{\scriptsize{\url{https://github.com/VisionLearningGroup/SSDA_MME}}}, trained using the same training policy and the few-shot labels specified by \cite{saito2019semi}.

Table~\ref{table:ssda-domainnet} illustrates the results of SSDA with various architectures, where S+T indicates the baseline few-shot learning algorithm based on cosine similarity learning~\cite{chen2019closerfewshot}. 
Our method consistently improves the performance with considerable margins in most adaptation tasks with respect to all tested architectures and baseline methods, which verifies its scalability to various conditions.

%% file: table/dg_pacs.tex
\begin{table}[t]
\caption{Multi-source domain generalization accuracy (\%) on PACS.
Each column title indicates the target domain.
The results of our DeepAll baselines \hs{(standard supervised learning on the mixture of source domains)} and SagNets are averaged over three repetitions.
SagNet$^{-\textrm{CBL}}$ and SagNet$^{-\textrm{ASBL}}$ refer to SagNets without content-biased learning and adversarial style-biased learning, respectively.
\hs{Our approach outperforms all competing methods, where each component of our method contributes to the performance improvements.}}
\begin{center}
\vspace{-4mm}
\scalebox{0.85}{
\begin{tabular}{lcccc|c}
\hline
 & Art paint. & Cartoon &  Sketch & Photo & Avg. \\
\hline
\multicolumn{6}{c}{AlexNet} \\
\hline
D-SAM & 63.87 & 70.70 & 64.66 & 85.55 & 71.20 \\
JiGen & 67.63 & 71.71 & 65.18 & 89.00 & 73.38 \\ 
Epi-FCR & 64.7 & 72.3 & 65.0 & 86.1 & 72.0 \\ 
MASF & {70.35} & {72.46} & {67.33} & \textbf{90.68} & {75.21} \\ 
MMLD & 69.27 & \textbf{72.83} & 66.44 & 88.98 & 74.38 \\ 
\hline
DeepAll & 65.19 & 67.83 & 63.75 & {90.08} & 71.71 \\ 
SagNet & \textbf{71.01} & 70.78 & \textbf{70.26} & 90.04 & \textbf{75.52} \\ 
\hline
\multicolumn{6}{c}{ResNet-18} \\
\hline
D-SAM & 77.33 & 72.43 & \textbf{77.83} & 95.30 & 80.72 \\
JiGen & 79.42 & 75.25 & 71.35 & {96.03} & 80.51 \\
Epi-FCR & 82.1 & 77.0 & 73.0 & 93.9 & 81.5 \\ 
MASF & 80.29 & {77.17} & 71.69 & 94.99 & 81.04 \\
MMLD & 81.28 & 77.16 & 72.29 & \textbf{96.09} & 81.83 \\ 
\hline
DeepAll & 78.12 & 75.10 & 68.43 & 95.37 & 79.26 \\
SagNet$^{-\textrm{CBL}}$ & 78.86 & 77.05 & 73.28 & 95.43  & 81.15 \\
SagNet$^{-\textrm{ASBL}}$ & {82.94} & 76.73 & 74.74 & 95.07  & {82.37} \\
SagNet & \textbf{83.58} & \textbf{77.66} & {76.30} & 95.47  & \textbf{83.25} \\ 
\hline
\end{tabular}
}
\end{center}
\label{table:msdg}
\end{table}

%% file: table/dg_officehome.tex
\begin{table}[t]
\caption{Multi-source domain generalization accuracy (\%) on Office-Home with a ResNet-18 backbone.}
\begin{center}
\vspace{-4mm}
\scalebox{0.85}{
\begin{tabular}{lcccc|c}
\hline
& Art & Clipart &  Product & Real-World & Avg. \\
\hline
D-SAM & 58.03 & 44.37 & 69.22 & 71.45 & 60.77 \\
JiGen & 53.04 & \textbf{47.51} & \textbf{71.47} & 72.79 & 61.20 \\ 
\cline{1-6}
DeepAll & 58.51 & 41.44 & 70.06 & 73.28 & 60.82 \\ 
SagNet$^{-\textrm{CBL}}$ & 60.00 & 42.85 & 70.11 & 73.12 & 61.52 \\ 
SagNet$^{-\textrm{ASBL}}$ & 59.31 & 41.89 & 70.44 & \textbf{73.52} & 61.29 \\ 
SagNet & \textbf{60.20} & 45.38 & 70.42 & 73.38 & \textbf{62.34} \\ 
\hline
\end{tabular}
}
\vspace{-10mm}
\end{center}
\label{table:msdg-officehome}
\end{table}

%% file: table/dg_pacs_single.tex
\begin{table*}[t]
\caption{Single-source domain generalization accuracy (\%) on PACS averaged over three repetitions (A: Art Painting, C: Cartoon, S: Sketch, P: Photo).
}
\begin{center}
\vspace{-4mm}
\scalebox{0.85}{
\begin{tabular}{lcccccccccccc|c}
\hline
 & A$\rightarrow$C & A$\rightarrow$S & A$\rightarrow$P & C$\rightarrow$A & C$\rightarrow$S & C$\rightarrow$P & S$\rightarrow$A & S$\rightarrow$C & S$\rightarrow$P & P$\rightarrow$A & P$\rightarrow$C & P$\rightarrow$S & Avg.\\
\hline
ResNet-18 & 
62.3 & 49.0 & 95.2 & 
65.7 & 60.7 & 83.6 & 
28.0 & 54.5 & 35.6 & 
64.1 & 23.6 & 29.1 & 54.3 \\
JiGen & 
57.0 &	50.0 &	\textbf{96.1} &
65.3 &	65.9 &	85.5 &
26.6 &	41.1 &	42.8 &
62.4 &	27.2 &	35.5 & 54.6 \\ 
ADA & 
64.3  &	\textbf{58.5}  &	94.5  &
66.7  &	65.6 &	83.6 &
37.0  &	58.6  &	41.6  &
65.3 &	32.7  &	35.9 & 58.7  \\ 
SagNet & 
\textbf{67.1} & 56.8 & 95.7 & 
\textbf{72.1} & \textbf{69.2} & \textbf{85.7} & 
\textbf{41.1} & \textbf{62.9} & \textbf{46.2} & 
\textbf{69.8} & \textbf{35.1} & \textbf{40.7} & \textbf{61.9} \\ 
\hline
\end{tabular}
}
\end{center}
\label{table:ssdg}
\end{table*}

%% file: table/uda_officehome.tex
\begin{table*}[t]
\caption{Unsupervised domain adaptation accuracy (\%) on Office-Home with varying adaptation methods and their SagNet combinations (A: Art, C: Clipart, P: Product, R: Real-World).
SagNets consistently boost the performance when combined with various adaptation methods.
}
\begin{center}
\vspace{-4mm}
\scalebox{0.85}{
\begin{tabular}{l|c|cccccccccccc|c}
\hline
  Method & SagNet & A$\rightarrow$C & A$\rightarrow$P & A$\rightarrow$R & C$\rightarrow$A & C$\rightarrow$P & C$\rightarrow$R & P$\rightarrow$A & P$\rightarrow$C & P$\rightarrow$R & R$\rightarrow$A & R$\rightarrow$C & R$\rightarrow$P & Avg.\\
\hline
\multirow{2}{*}{ResNet-50} & & 41.3	&63.8	&71.4	&49.1	&59.6	&61.4	&46.8	&36.1	&68.8	&63.0	&45.9	&76.5	&57.0\\
 & \checkmark & \textbf{45.7}	&\textbf{64.1}	&\textbf{72.6}	&\textbf{49.6}	&\textbf{60.0}	&\textbf{63.5}	&\textbf{49.9}	&\textbf{40.7}	&\textbf{71.1}	&\textbf{64.8}	&\textbf{50.9}	&\textbf{78.1}	&
\textbf{59.2}\\
\hline
\multirow{2}{*}{DANN} & & 44.7	&62.7	&70.3	&47.1	&60.1	&61.4	&46.1	&41.7	&68.5	&62.3	&50.9	&76.7	&57.7\\
& \checkmark &\textbf{48.8}	&\textbf{65.2}	&\textbf{71.4}	&\textbf{50.3}	&\textbf{61.4}	&\textbf{62.5}	&\textbf{50.7}	&\textbf{45.7}	&\textbf{71.8}	&\textbf{65.4}	&\textbf{55.2}	&\textbf{78.6}	&\textbf{60.6}\\
\hline
\multirow{2}{*}{JAN}  &  & 
45.0    &63.3    &72.6    &
53.3    &\textbf{66.0}    &64.4    &
50.9    &40.8    &72.1    &
64.9    &49.4    &78.8    &60.1 \\
& \checkmark &
\textbf{50.1} 	&\textbf{66.8} 	&\textbf{73.9} 	&
\textbf{56.9}	&64.7	&\textbf{66.1} 	&
\textbf{54.9}	&\textbf{45.6}	&\textbf{75.2}	&
\textbf{70.0}	&\textbf{55.3}	&\textbf{80.1}	&\textbf{63.3}\\
\hline
\multirow{2}{*}{CDAN} & & 50.6	&69.0	&\textbf{74.9}	&54.6	&66.1	&67.9	&57.2	&46.9	&75.6	&69.1	&55.8	&\textbf{80.6}	&64.0\\
& \checkmark & \textbf{53.2}	&\textbf{69.2}	&\textbf{74.9}	&\textbf{55.9}	&\textbf{67.8}	&\textbf{68.6}	&\textbf{58.1}	&\textbf{51.8}	&\textbf{76.4}	&\textbf{69.8}	&\textbf{58.1}	&80.4	&\textbf{65.3}\\
\hline
\multirow{2}{*}{SymNet}  &  & 
47.2     &70.9     &77.4     &
64.5     &\textbf{71.7}     &73.0     &
63.3     &49.0     &\textbf{79.1}     &
74.2     &54.1     &82.7     &67.3  \\
& \checkmark &
\textbf{49.6} 	&\textbf{72.4} 	&\textbf{77.9} 	&
\textbf{63.9}	&71.1  	        &\textbf{72.5} 	&
\textbf{65.2}	&\textbf{51.0}	&\textbf{79.1}	&
\textbf{74.9}	&\textbf{56.1}	&\textbf{83.0}	&\textbf{68.0}\\
\hline
\end{tabular}
}
\end{center}
\vspace{-2mm}
\label{table:uda-officehome}
\end{table*}

%% file: table/uda_domainnet_v2.tex
\begin{table*}[t]
\caption{Unsupervised domain adaptation accuracy (\%) on DomainNet (C: Clipart, P: Painting, S: Sketch, R: Real).
SagNets improve the performance over both ResNet-18 and CDAN~\cite{mingsheng2018conditional} baselines.
}
\begin{center}
\vspace{-4mm}
\scalebox{0.85}{
\begin{tabular}{l|c|ccccccc|c}
\hline
Method & SagNet & R$\rightarrow$C &  R$\rightarrow$P &  P$\rightarrow$C &  C$\rightarrow$S & S$\rightarrow$P & R$\rightarrow$S & P$\rightarrow$R &Avg.\\

\hline
 \multirow{2}{*}{ResNet-18} & & 53.1	&57.7	&52.4	&47.5	&52.0	&43.4	&\textbf{68.5}	&53.5\\
&   \checkmark & \textbf{54.4}	&\textbf{58.0}	&\textbf{53.1}	&\textbf{49.2}	&\textbf{52.2}	&\textbf{46.4}	&67.4	&\textbf{54.4}\\
\hline
 \multirow{2}{*}{CDAN} & & 53.0	&57.4	&52.3	&48.0	&52.3	&43.4	&\textbf{67.2}	&53.4\\
&  \checkmark&\textbf{54.4}	&\textbf{59.4}	&\textbf{52.8}	&\textbf{49.5}	&\textbf{52.4}	&\textbf{45.9}	&67.1	&\textbf{54.5}\\
\hline
\end{tabular}
}
\end{center}
\label{table:uda-domainnet}
\end{table*}

%% file: table/ssda_domainnet2.tex
\begin{table*}[t]
\caption{Semi-supervised domain adaptation accuracy (\%) on DomainNet (C: Clipart, P: Painting, S: Sketch, R: Real).
SagNets consistently improve the performance over various baselines in terms of backbone architectures and adaptation methods.
}
\begin{center}
\vspace{-4mm}
\scalebox{0.78}{
\begin{tabular}{l | l | c | cccccccccccccc|cc }
\hline
\multirow{2}{*}{Backbone} 
& \multirow{2}{*}{Method} 
& \multirow{2}{*}{SagNet} 
&\multicolumn{2}{c}{R $\rightarrow$ C}
&\multicolumn{2}{c}{R $\rightarrow$ P} 
& \multicolumn{2}{c}{P $\rightarrow$ C}  
& \multicolumn{2}{c}{C $\rightarrow$ S} 
& \multicolumn{2}{c}{S $\rightarrow$ P} 
& \multicolumn{2}{c}{R $\rightarrow$ S} 
& \multicolumn{2}{c|}{P $\rightarrow$ R}   
&\multicolumn{2}{c}{Avg.} \\ 
& &  & 1\scriptsize{-shot}&3\scriptsize{-shot} &1\scriptsize{-shot}&3\scriptsize{-shot}&1\scriptsize{-shot}&3\scriptsize{-shot} &1\scriptsize{-shot}&3\scriptsize{-shot}&1\scriptsize{-shot}&3\scriptsize{-shot}&1\scriptsize{-shot}&3\scriptsize{-shot} &1\scriptsize{-shot}&3\scriptsize{-shot} &1\scriptsize{-shot}&3\scriptsize{-shot}  \\ \hline

 \multirow{4}{*}{AlexNet}
&\multirow{2}{*}{S+T} & & 43.3 & 47.1 & 42.4 & 45.0 & 40.1 & 44.9 & 33.6 & 36.4 & 35.7 & 38.4 & 29.1 & 33.3 & \textbf{55.8} & \textbf{58.7} & 40.0 & 43.4 \\
& & \checkmark &\textbf{45.8}	&\textbf{49.1}	&\textbf{45.6}	&\textbf{46.7}	&\textbf{42.7}	&\textbf{46.3}	&\textbf{36.1}	&\textbf{39.4}	&\textbf{37.1}	&\textbf{39.8}	&\textbf{34.2}	&\textbf{37.5}	&54.0	&57.0	&\textbf{42.2}	&\textbf{45.1}\\
\cline{2-19}
& \multirow{2}{*}{MME} & & 48.9  &  55.6 &  48.0 &  49.0 &  46.7 & 51.7 &  36.3 &  39.4 &  39.4 &  43.0 &  33.3 &  37.9 &  56.8 &  \textbf{60.7} &  44.2 &  48.2  \\
& & \checkmark & \textbf{54.1}  & \textbf{58.6}  & \textbf{49.8} & \textbf{52.2} & \textbf{48.7} & \textbf{54.4} & \textbf{39.6} & \textbf{43.4} & \textbf{40.4} & \textbf{43.4 }& \textbf{39.7} & \textbf{42.8} & \textbf{57.0} & 60.3 & \textbf{47.0} & \textbf{50.7}  \\
\hline
\hline
\multirow{4}{*}{VGG-16}
& \multirow{2}{*}{S+T} & & 49.0 & 52.3 & 55.4 & 56.7 & 47.7 & 51.0 & 43.9 & 48.5 & 50.8 & 55.1 & 37.9 & 45.0 & \textbf{69.0} & \textbf{71.7}  & 50.5 & 54.3 \\
& & \checkmark &\textbf{51.8}	&\textbf{54.9}	&\textbf{57.8}	&\textbf{59.4}	&\textbf{50.4}	&\textbf{54.2}	&\textbf{48.9}	&\textbf{52.9}	&\textbf{53.1}	&\textbf{56.3}	&\textbf{45.6}	&\textbf{49.4}	&68.3	&70.9	&\textbf{53.7}	&\textbf{56.9}\\
\cline{2-19}
 & \multirow{2}{*}{MME} & &  60.6 &  64.1 &  63.3 &  63.5 &  57.0 &  60.7 & 50.9 &  55.4 &  \textbf{60.5} &  60.9 &  50.2 &  54.8 &  \textbf{72.2} &  \textbf{75.3}  &  59.2 &  62.1 \\
& & \checkmark & \textbf{64.9} & \textbf{67.8} & \textbf{64.5} & \textbf{66.0}  & \textbf{60.4} & \textbf{65.8} & \textbf{54.7} & \textbf{59.0} & 59.8 & \textbf{62.0} & \textbf{56.6} & \textbf{59.6} & 71.1 & 74.2 & \textbf{61.7} & \textbf{64.9} \\
\hline
\hline 
\multirow{4}{*}{ResNet-34} & 
\multirow{2}{*}{S+T}  & & 55.6 & 60.0 & 60.6 & 62.2 & 56.8 & 59.4 & 50.8 & 55.0 & 56.0 & \textbf{59.5} & 46.3 & 50.1 & 71.8 & \textbf{73.9} & 56.9 & 60.0 \\
& &\checkmark & \textbf{59.4  }& \textbf{62.0}& \textbf{61.9} & \textbf{62.9  }&\textbf{59.1} &\textbf{61.5} & \textbf{54.0} & \textbf{57.1} & \textbf{56.6} &59.0 &  \textbf{49.7}&\textbf{54.4}  &\textbf{72.2}  & 73.4 & \textbf{59.0} & \textbf{61.5}  \\
\cline{2-19}
&\multirow{2}{*}{MME} & & 70.0 &  72.2 &  67.7 &  69.7 &  69.0 &  71.7 &  56.3 &  61.8 &  64.8 &  66.8 &  61.0 &  61.9 &  \textbf{76.1} & \textbf{78.5} &  66.4 &  68.9  \\
& & \checkmark & \textbf{72.3}  & \textbf{74.2}& \textbf{69.0} & \textbf{70.5 }&\textbf{70.8 }& \textbf{73.2} & \textbf{61.7} &\textbf{64.6}  & \textbf{66.9} &\textbf{68.3}  & \textbf{64.3} &  \textbf{66.1}& 75.3 & 78.4 & \textbf{68.6} & \textbf{70.8}  \\
\hline
\end{tabular}
}
\end{center}
\vspace{-2mm}
\label{table:ssda-domainnet}
\end{table*}

%% file: 5.conclusion.tex
\section{Conclusion}
\label{sec:conclusion}
\cm{
We present Style-Agnostic Networks (SagNets) that are robust against domain shift caused by style variability across domains.
By randomizing styles in a latent feature space and adversarially disentangling styles from class categories, SagNets are trained to rely more on contents rather than styles in their decision making process.
Our extensive experiments confirm the effectiveness of SagNets in controlling the inductive biases of CNNs and reducing domain discrepancy in a broad range of problem settings.
}

Our work is orthogonal to many existing domain adaptation approaches that explicitly align the distributions of different domains, and achieves further improvements by complementing those methods.
The principle of how we deal with the intrinsic biases of CNNs can also be applied to other areas such as improving robustness under image corruptions~\cite{hendrycks2019robustness} and defending against adversarial attacks~\cite{goodfellow2015explaining}, which are left for future investigation.